\documentclass{WileyMSP-template}

\usepackage{array}
\usepackage{booktabs}
\usepackage[colorlinks=true, linkcolor=blue, citecolor=blue, urlcolor=blue]{hyperref}
\usepackage[table]{xcolor}

\definecolor{tableblue}{HTML}{DCEAF7}
\definecolor{tablegray}{HTML}{F7F8FA}
\definecolor{tableline}{HTML}{98A2B3}

\begin{document}

\pagestyle{fancy}
\rhead{\includegraphics[width=2.5cm]{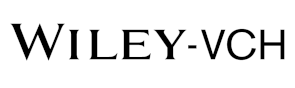}}

\title{A Practical Noise2Noise Denoising Pipeline for High-Throughput Raman Spectroscopy}

\maketitle


\author{David Martin-Calle*,}
\author{Cesar Alvarez-Llamas,}
\author{Vincent Motto-Ros,}
\author{Christophe Dujardin,}
\author{Jérémie Margueritat*,}
\author{David Rodney*}

\begin{affiliations}
D. Martin-Calle\\
Université Claude Bernard Lyon 1, CNRS, Institut Lumière Matière, F-69622 Villeurbanne, France\\
Email Address: david.martin-calle@univ-lyon1.fr

C. Alvarez-Llamas\\
Université Claude Bernard Lyon 1, CNRS, Institut Lumière Matière, F-69622 Villeurbanne, France

V. Motto-Ros\\
Université Claude Bernard Lyon 1, CNRS, Institut Lumière Matière, F-69622 Villeurbanne, France

C. Dujardin\\
Université Claude Bernard Lyon 1, CNRS, Institut Lumière Matière, F-69622 Villeurbanne, France\\
Institut Universitaire de France (IUF), Paris, France

J. Margueritat\\
Université Claude Bernard Lyon 1, CNRS, Institut Lumière Matière, F-69622 Villeurbanne, France\\
Email Address: jeremie.margueritat@univ-lyon1.fr

D. Rodney\\
Université Claude Bernard Lyon 1, CNRS, Institut Lumière Matière, F-69622 Villeurbanne, France\\
Email Address: david.rodney@univ-lyon1.fr

\end{affiliations}




\begin{abstract}

A lightweight and reproducible denoising pipeline for high-throughput Raman spectroscopy is presented. The approach relies on a one-dimensional convolutional autoencoder trained using a Noise2Noise strategy, requiring neither external spectral libraries nor high signal-to-noise reference spectra for training. From a reduced training subset composed of repeated short-exposure acquisitions, the model learns to reconstruct Raman spectra while efficiently suppressing stochastic noise. The method is evaluated on a heterogeneous mineral sample, using both quantitative spectral fidelity metrics (RMSE, SNR, SSIM) and task-oriented criteria based on unsupervised K-means classification. Results demonstrate that integration times as short as 5 ms per spectrum, which are typically insufficient for reliable interpretation, yield denoised spectra with high fidelity to the reference data while preserving chemically coherent maps. This work provides a practical trade-off between spectral quality and acquisition speed, enabling fast, adaptable Raman workflows compatible with routine laboratory use. It also offers a transferable framework for other one-dimensional spectroscopic modalities.

\end{abstract}

\section{Introduction}

Raman spectroscopy is a well-established laser-based technique to characterize the chemical composition and crystallographic structure of materials with microscopic spatial resolution. Its sensitivity to local bonding environments and molecular structure has made it a key analytical tool across diverse fields, including materials science, chemistry, and biology~\cite{Ferraro2003, Smith2019, Orlando2021}.
When mapping extended areas, a full Raman spectrum is acquired at each point of a scanned region to construct a hyperspectral map, a spatially resolved dataset that reveals local chemical and structural heterogeneities within the sample~\cite{Delhaye1975, Guhlke2016, Trevisan2012}. However, Raman scattering is intrinsically weak, with typically only one in $10^{6}$--$10^{8}$ incident photons undergoing inelastic scattering~\cite{Long2002, Jahn2021}. As a result, achieving spectra with high signal-to-noise ratio (SNR) often requires long acquisition times, particularly for weakly scattering materials. This trade-off fundamentally limits the throughput of high-resolution or large-area Raman experiments, constraining its use in time-sensitive or high-throughput workflows~\cite{Zhao2022a, Schlucker2014}.

Several experimental strategies have been developed to overcome the inherently weak Raman signal. Increasing the laser power can enhance the signal intensity but risks sample heating, photodegradation, or unwanted fluorescence. Resonance Raman spectroscopy exploits resonance with electronic transitions to selectively enhance specific vibrational modes, but is applicable only to materials with suitable absorption bands~\cite{Long2002, Smith2019}. Surface-Enhanced and Tip-Enhanced Raman Spectroscopy (SERS and TERS) utilize localized surface plasmon resonances in metallic nanostructures to amplify the electromagnetic field by several orders of magnitude, up to $10^{6}$--$10^{8}$, enabling single-molecule sensitivity under optimized conditions~\cite{Schlucker2014, Pettinger2012, LeRu2009}. However, both techniques require precisely engineered substrates or tips and are highly sensitive to surface cleanliness, nanostructure geometry, and optical alignment, which limits their reproducibility and applicability for routine or large-area analyses~\cite{LeRu2009, Schlucker2014}.

Despite these advances, post-acquisition denoising remains essential to achieve practical mapping times without compromising spectral quality. Traditional numerical approaches include moving-window smoothing methods such as the Savitzky-Golay filter~\cite{Savitzky1964} and frequency-domain filtering using Fourier or wavelet transforms~\cite{Mallat1999}. More recently, deep learning-based methods, including convolutional neural networks (CNNs), denoising autoencoders (DAEs), and U-Nets, have demonstrated superior performance by learning to reconstruct clean spectra from noisy inputs~\cite{Fang2024, Boateng2025}. However, current methods vary widely in computational efficiency and portability, which can limit their suitability for high-throughput Raman denoising and mapping~\cite{Luo2022}.

Recent self-supervised Raman denoising studies differ in signal structure and target application. Wu et al.\ \cite{Wu2024} addressed generic Raman and SERS spectrum denoising with a self-supervised U-Net, and reported moderate SNR improvements on the order of $3$--$5\times$. Jiang et al.\ \cite{Jiang2025} proposed a 3D Noise2Void strategy for multidimensional Raman hyperspectral imaging, and Chen et al.\ \cite{Chen2026} developed a specialized method for single-particle Raman analysis in a mixed Gaussian -- flicker noise environment. In contrast to image-based Raman denoising methods \cite{Kumari2024}, the present work focuses on repeated short-exposure one-dimensional Raman spectra and targets high-throughput analysis of static samples, with evaluation at the workflow level including training, validation, and inference overheads.

More broadly, Noise2Noise-type training is relevant beyond Raman spectroscopy itself. Lehtinen et al.\ \cite{Lehtinen2018} introduced the general noisy-target denoising principle, while Luo et al.\ \cite{Luo2024S2S} and Karpov et al.\ \cite{Karpov2026} extended related ideas to chemical hyperspectral imaging and X-ray fluorescence microscopy. These studies support the broader applicability of self-supervised denoising in chemically resolved measurements, while the present work targets the specific case of repeated short-exposure one-dimensional Raman spectra for high-throughput analysis of static samples. We expect that similar strategies will also be relevant to other spectroscopic modalities or experimental setups when comparable repeated-acquisition data are available.

Here, we present a practical Noise2Noise-based denoising framework for high-throughput Raman spectroscopy. The contribution of this work is threefold: (i) a complete and ready-to-use denoising pipeline released as open-source code; (ii) a lightweight one-dimensional convolutional autoencoder that is easy to configure and use, and that can be trained and deployed under routine experimental conditions without large external datasets or specialized hardware; and (iii) substantial denoising performance under highly noise-limited acquisition conditions, with SNR improvements exceeding two orders of magnitude at 5~ms while maintaining spectral fidelity and mapping accuracy.

The complete production pipeline code is provided on Zenodo \href{https://doi.org/10.5281/zenodo.18154207}{10.5281/zenodo.18154207}~\cite{Martin-Calle2026}.

\section{Experimental sample and Raman spectra acquisition}

\subsection{Sample}

\begin{figure}[h!]
    \centering
    \includegraphics[width=0.75\linewidth]{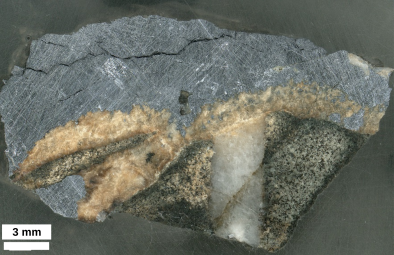}
    \caption{Optical image of the mineral sample used to benchmark the proposed denoising approach (see text for details).}
    \label{fig:sample}
\end{figure}

To demonstrate the potential of our denoising strategy, we selected the mineral sample shown in \autoref{fig:sample} from the polymetallic W--Au--Pb--Zn--Ag (Sb--Ba) district of Tighza, Morocco, a well-known mining area characterized by a complex association of metallic elements. This polymetallic composition results from successive hydrothermal mineralization processes, which led to the coexistence of several metal-bearing phases within the same geological environment. The Tighza district has historically been exploited for lead and silver and is documented as a reference site for multi-metal mineralization~\cite{Bouabdellah2016}.

The studied sample is associated with the Sidi Ahmed hydrothermal event, a mineralizing episode responsible for part of the polymetallic enrichment in the district. As a consequence of this geological history, the sample exhibits strong chemical and mineralogical heterogeneity at the microscale, making it a particularly challenging case for spectral analysis. In addition, this sample has previously been characterized in our laboratory using Laser-Induced Breakdown Spectroscopy (LIBS) imaging, providing an independent estimation of the dominant mineral phases present \cite{Nardecchia2020}, which include quartz, galena, ankerite and some aluminosilicates. This prior knowledge makes the sample well suited for assessing the performance of the proposed denoising approach.

The analyzed rock section measures approximately 3.2 cm $\times$ 1.6 cm, with a thickness of about 1 cm. Prior to Raman imaging, the sample was embedded in resin and subsequently carefully polished to ensure optimal optical quality and reproducibility of the measurements.

\subsection{Raman spectra acquisition}
\label{subsec:acquisition}

\begin{table}[h!]
\centering
\small
\begin{minipage}{0.52\linewidth}
\centering
\rowcolors{3}{tablegray}{white}
\setlength{\tabcolsep}{6pt}
\renewcommand{\arraystretch}{1.15}
\begin{tabular}{>{\bfseries}l !{\color{tableline}\vrule width 0.8pt} c !{\color{tableline}\vrule width 0.8pt} c}
\toprule
\rowcolor{tableblue}
\textbf{Parameter}  & \textbf{Final map} & \textbf{Train subset} \\
\midrule
Acquisition time (ms) & \multicolumn{2}{c}{$5$, $10$, $25$, $50$, $100$} \\
Laser power & \multicolumn{2}{c}{$25$ mW} \\
Laser wavelength & \multicolumn{2}{c}{$532$ nm} \\
Spectral resolution & \multicolumn{2}{c}{$1.97$ - $2.39$ cm$^{-1}$} \\
Spectral range & \multicolumn{2}{c}{$59.9$ - $1791.5$ cm$^{-1}$} \\
Grid size (points) & $320 \times 200$ & $32 \times 22$ \\
Step size (spatial resolution) & $85\ \mu$m & $650\ \mu$m \\
Repetitions & $1$ & \textit{see adjacent} \\
\bottomrule
\end{tabular}
\end{minipage}%
\hspace{0.05\linewidth} 
\begin{minipage}{0.35\linewidth}
\centering
\textbf{TRAIN SUBSET}

\vspace{0.35em}

\rowcolors{3}{tablegray}{white}
\setlength{\tabcolsep}{6pt}
\renewcommand{\arraystretch}{1.15}
\begin{tabular}{>{\bfseries}c !{\color{tableline}\vrule width 0.8pt} c}
\toprule
\rowcolor{tableblue}
\textbf{Acquisition time} & \textbf{Repetitions} \\
\midrule
$100$ ms & $150$ \\
$50$ ms & $200$ \\
$25$ ms & $300$ \\
$10$ ms & $1000$ \\
$5$ ms & $1500$ \\
\bottomrule
\end{tabular}
\end{minipage}
\caption{Raman acquisition parameters and repetition scheme for the training subset and final map.}
\label{table:acquisition}
\end{table}

The heterogeneous structure of the studied sample produces distinct Raman signatures, enabling the evaluation of denoising and clustering performance across multiple mineral phases. Experimental measurements were carried out on the LIBELUL platform at the Institut Lumière Matière (iLM) in Lyon. A 532~nm Cobolt laser \cite{Hardy2025} was used as the excitation source. It was focused onto the sample using a 10× objective (numerical aperture NA = 0.28, working distance = 34 mm, focal length = 20 mm). The scattered light was collected by the same objective in a backscattering geometry and directed to a Czerny–Turner spectrograph coupled to an electron-multiplying charge-coupled device (EMCCD) detector, providing high sensitivity for low-intensity Raman signals. The laser power was carefully adjusted to optimize the signal-to-noise ratio while minimizing the risk of sample heating or surface damage \cite{Vulchi2024}. The camera was cooled to $-80\,^{\circ}\mathrm{C}$ to minimize thermal noise.

The analytical protocol began with the acquisition of a training dataset. The number of measurement points and the lateral resolution (step size) were chosen to balance total acquisition time with sufficient sampling of the distinct phases identified by optical microscopy, resulting in a grid of $32 \times 22 = 704$ spatial points. To obtain the cleanest possible ground-truth spectra after preprocessing (see \autoref{subsec:preproc}), multiple repeated acquisitions were performed at several integration times, ranging from short 5 ms acquisitions repeated 1500 times to long 100 ms acquisitions repeated 150 times (see the right-hand side of \autoref{table:acquisition}).

Following acquisition of the training dataset, the final Raman maps were recorded immediately over the same sample area using a finer spatial grid, yielding a higher lateral resolution. The mapping coordinates were aligned with those of the training acquisition, within the limits imposed by stage positioning uncertainty. This procedure ensures that the sparsely sampled training data remain spatially representative of the high-resolution Raman maps.

All raw Raman spectra acquired at different integration times have been deposited in an open-access Zenodo repository \href{https://doi.org/10.5281/zenodo.18244161}{10.5281/zenodo.18244161} \cite{Alvarez-Llamas2026}.

\subsection{Noise composition}
\label{subsec:noise}

\begin{figure}[h!]
    \centering
    \includegraphics[width=0.6\linewidth]{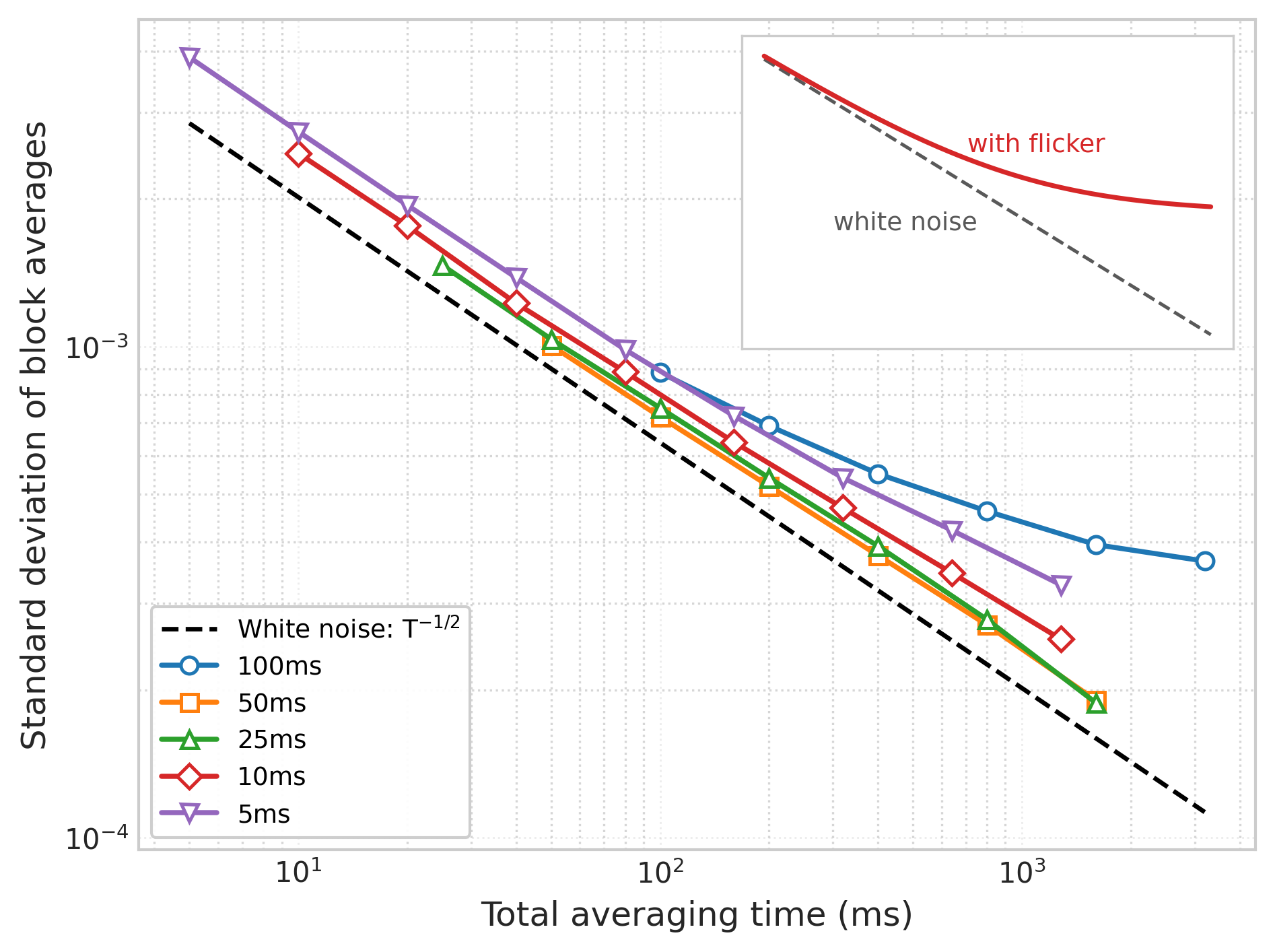}
    \caption{Block-averaging analysis of Raman noise for acquisition times from 5 to 100 ms as a function of total averaging time. The dashed black line shows the $T^{-1/2}$ scaling expected for white noise. A clear departure from this behavior is observed only for the 100 ms data, indicating that low-frequency correlations become significant only at this acquisition time. The inset schematically illustrates the expected departure from white-noise averaging in the presence of flicker-like noise.}
    \label{fig:flicker}
\end{figure}

Understanding the composition of the noise is essential to validate the denoising approach proposed in this work. Raman noise combines several physical and instrumental contributions with different statistical properties \cite{Smulko2015}. Photon (shot) noise follows Poisson statistics and its variance scales with signal intensity, so its importance increases at longer acquisition times relative to detector read noise. By contrast, detector read noise is reasonably described as signal-independent Gaussian noise and becomes relatively more important at shorter acquisition times, when the collected Raman signal is weaker. Thermal (dark) noise also follows Poisson statistics, but its contribution is reduced here by cooling the camera to $-80\,^{\circ}\mathrm{C}$. Taken together, these components can be treated, to a good approximation, as centered around the underlying signal and independent between repeated acquisitions of the same spatial point, even though their variance may change across the spectrum.

The block-averaging analysis in \autoref{fig:flicker} nevertheless indicates that this approximation is not exact at the longest integration time. For the \(100\,\mathrm{ms}\) measurements, the noise decreases more slowly than the \(T^{-1/2}\) law expected for purely white noise, which points to an additional low-frequency correlated contribution. This behavior is consistent with flicker-like (\(1/f\)) noise, which does not average down indefinitely and can produce a flicker floor at long averaging times \cite{Vernotte2015}. By contrast, for shorter integration times the observed scaling remains close to the white-noise regime, suggesting that correlations between repetitions are weak enough to preserve the practical validity of the independent-noise assumption used later for Noise2Noise training. The robustness of the Noise2Noise approach to approximate rather than exact noise-statistics assumptions is further discussed in \autoref{subsubsec:noise2noise}.
\section{Denoising method}

Prior to training and denoising, all Raman spectra were subjected to a standardized preprocessing workflow summarized in \autoref{fig:flowchart-preprocess} and detailed in the section below.

\subsection{Data preprocessing}
\label{subsec:preproc}

\begin{figure}[h!]
    \centering
    \includegraphics[width=\linewidth]{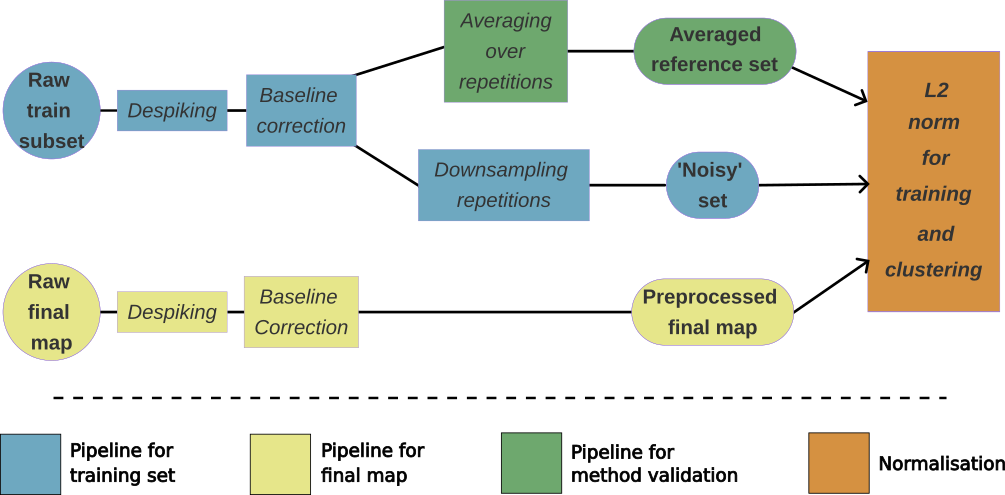}
    \caption{Schematic overview of the preprocessing workflow applied to the Raman spectra. Raw measurements undergo despiking and baseline correction before being partitioned into an ``averaged reference'' dataset, a ``noisy'' dataset, and a production dataset. These processed spectra are then normalized and used respectively for evaluation, model training, and final map reconstruction.}
    \label{fig:flowchart-preprocess}
\end{figure}

Preprocessing begins by cropping both ends of all spectra to remove edge artefacts, including residual Rayleigh scattering at low Raman shifts. The spectral resolution is then reduced to 736 Raman shift values in order to standardize the input dimensionality across all datasets. The raw spectra used for training and validation are subsequently processed to generate two datasets, referred to as the ``averaged reference'' and ``noisy'' sets (\autoref{fig:flowchart-preprocess}).

Each spectrum is first despiked to remove cosmic-ray artefacts \cite{Vulchi2024}. Spikes are identified as narrow positive peaks with a width below three pixels and an amplitude exceeding five times the local standard deviation, computed within a sliding window of 40 pixels excluding the peak itself \cite{Coca-Lopez2024}. Detected spikes are replaced by linear interpolation, yielding a cleaned spectrum. This procedure is applied iteratively to each spectrum until no additional spikes are detected. Despiking is followed by polynomial baseline correction \cite{Liu2015, Hu2018} using a fourth-degree fit, in order to compensate for fluorescence contributions and slowly varying background trends. At this point, the preprocessing workflow splits into two parallel branches.

The first branch (green pathway in \autoref{fig:flowchart-preprocess}) retains the full set of repeated measurements and computes an averaged spectrum for each spatial point. The second branch (blue pathway in \autoref{fig:flowchart-preprocess}) performs a controlled downsampling by retaining only a fixed number of spectra acquired with a single integration time, thereby emulating realistic acquisition constraints typical of high-throughput or production workflows. In the present study, 20 repetitions per point are retained, corresponding to an effective acquisition time of approximately 22~\% of the total map acquisition time.

Prior to both training and clustering, the noisy spectra are L2-normalized to enforce a consistent intensity scale across the dataset. Spectral normalization is commonly employed in machine-learning pipelines \cite{Lasch2012, Gautam2015} to mitigate the influence of global intensity variations arising from experimental factors such as laser power fluctuations, focus variations, or local differences in scattering efficiency \cite{Luo2022}. After denoising, only spectra intended for visualization or physical interpretation are rescaled to their original norms, whereas spectra used for clustering remain L2-normalized. This strategy preserves physically meaningful intensity information in the final outputs while maintaining a normalized feature space for robust classification.

The ``noisy'' dataset is used to train the denoising model (\autoref{subsec:training}), while the ``averaged reference'' dataset serves as a reference for evaluation and metric-based validation (\autoref{subsec:metrics}).

A third preprocessing workflow, shown in yellow in \autoref{fig:flowchart-preprocess}, applies the same sequence of despiking, baseline correction, and L2 normalization. This workflow is used for production data prior to denoising.

\subsection{Denoising network architecture}

\begin{figure}[h!]
    \centering
    \includegraphics[width=\linewidth]{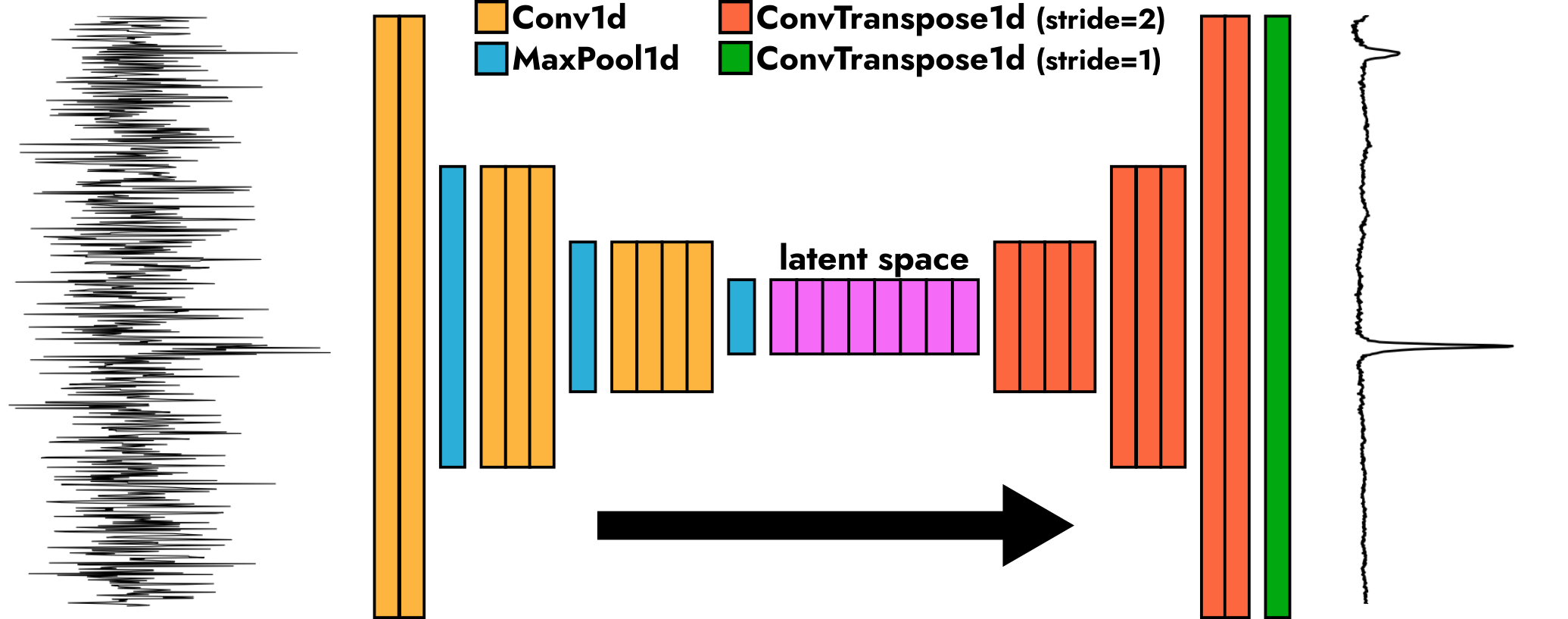}
    \caption{Simplified schematic of the 1D convolutional autoencoder used for Raman spectral denoising. The network compresses noisy spectra into a latent representation and reconstructs a denoised output, preserving peak features while attenuating stochastic noise. Left: example of a noisy input spectrum acquired with a 5~ms integration time. Right: corresponding denoised output spectrum. Adjacent blocks denote the progressive increase in the number of filters, which allows the network to extract increasingly rich spectral features and better distinguish Raman peaks from noise. The detailed architecture and hyperparameters are presented in \autoref{subsubsec:architecture}.}
    \label{fig:denoiser}
\end{figure}

The denoising stage employs a fully convolutional one-dimensional autoencoder specifically tailored for Raman spectral data (\autoref{fig:denoiser}). 

Fully convolutional denoising autoencoders are particularly well suited to Raman spectral data, as they preserve local spectral correlations and translational invariance while significantly reducing the number of trainable parameters and the risk of overfitting. As a result, they achieve more effective suppression of stochastic noise while better preserving peak shape, position, and relative intensity compared with dense-layer architectures \cite{Vincent2008, Chiang2019}.

This class of neural network encodes a noisy input spectrum into a compressed latent representation that preserves essential structural information while attenuating noise and artefacts \cite{Vincent2008}. During the subsequent decoding phase, the network reconstructs a clean approximation of the original Raman signal. By limiting the representational capacity of the latent space, the model is encouraged to suppress non-informative variations and enhance diagnostically relevant features, such as peak shape and position \cite{Vincent2008, LeCun2015}.

\subsubsection{Detailed architecture}
\label{subsubsec:architecture}

The one-dimensional convolutional autoencoder employed in this work is organized into 11 convolutional blocks: five encoder blocks, one latent block, and five decoder blocks, followed by a final one-channel ConvTranspose1d output projection. The role of the encoder is to progressively transform the input Raman spectrum into a compact representation that emphasizes meaningful spectral features while reducing noise. This is achieved by gradually increasing the number of convolutional filters across successive layers ([16, 24, 32, 48, 64]), allowing the network to capture increasingly complex spectral patterns. After each convolutional block, MaxPool1d layers reduce the spectral resolution by locally retaining only the most prominent values within small spectral windows, effectively compressing the data while suppressing high-frequency noise. In the implementation used here, each encoder convolution is applied with stride 1, while each MaxPool1d layer uses a pooling kernel of size 2 with stride 2.

At the center of the network, a latent layer with 96 filters provides a compact yet expressive representation of the spectrum. This latent space is designed to retain the essential characteristics of Raman peaks, such as their position, shape, and relative intensity, while discarding non-informative fluctuations. All Conv1d and ConvTranspose1d layers use a kernel size of 11, chosen to match the typical width of Raman peaks, which extend over several neighbouring spectral points. By covering an entire peak within a single convolutional window, the network can model peak shapes directly rather than relying on isolated intensity values, while still suppressing rapid variations dominated by noise \cite{Chin2025, Soysal2025}.

The decoder mirrors the encoder structure, with a symmetric reduction in the number of filters ([64, 48, 32, 24, 16]) and ConvTranspose1d layers with stride 2 that progressively restore the original spectral resolution by expanding the compressed representation back to its initial length. This stage reconstructs a denoised spectrum from the latent representation. A final one-channel ConvTranspose1d output layer with stride 1 then produces the reconstructed Raman spectrum at full resolution.

Rectified linear unit (ReLU) activation functions \cite{Nair2010, Glorot2011} are used in all hidden layers to introduce non-linearity while avoiding saturation effects that can hinder training. No activation function is applied in the output layer, since Raman intensities are continuous physical quantities. A linear output allows the network to reproduce the full dynamic range of spectral intensities and avoids artificial clipping or distortion of peak amplitudes that could arise from bounded activation functions \cite{Hinton2006, Vincent2010}.

The final configuration contains 262{,}705 trainable parameters, corresponding to 1.05~MB of storage in float32 precision. This compactness, and the resulting computational efficiency, is an important design feature because training time becomes a practical constraint when denoising spectra acquired with integration times as short as 5~ms. The proposed model is compact compared with representative recent architectures reported for Raman denoising. For clarity, a comparison with representative recent models is provided in the following table.

\begin{table}[h!]
\centering
\rowcolors{2}{tablegray}{white}
\setlength{\tabcolsep}{6pt}
\renewcommand{\arraystretch}{1.15}
\begin{tabular}{>{\bfseries}l !{\color{tableline}\vrule width 0.8pt} c c}
\toprule
\rowcolor{tableblue}
Model / reference & Size (MB) & Approx.\ parameters \\
\midrule
This work & 1.05 & 262{,}705 \\
Wu et al.\ \cite{Wu2024} & 1.76 & $\sim$440{,}000 \\
Horgan et al.\ \cite{Horgan2021} & 8.34 & $\sim$2{,}000{,}000 \\
\bottomrule
\end{tabular}
\end{table}

As shown, the model reported by Wu et al.\ \cite{Wu2024} -- already described by its authors as compact relative to a conventional CNN -- remains significantly larger than the present architecture. The Res U-Net proposed by Horgan et al.\ \cite{Horgan2021} is larger still by a substantial margin. Taken together, these comparisons support characterizing the present architecture as lightweight in the context of Raman spectral denoising. This point is also supported by the measured runtimes reported later in \autoref{tab:computational-cost}: training requires only 20.31~s on the laptop workstation used in this study, and inference on the full final map of 64{,}000 spectra requires only 1.12~s.

The network architecture and associated hyperparameters were selected through an automated optimization procedure based on the Tree-structured Parzen Estimator (TPE) algorithm \cite{Bergstra2011}. This approach explores the hyperparameter space efficiently and enabled the identification of a configuration that provides robust denoising performance across a range of training conditions, including variations in noise level and acquisition parameters.

\subsection{Training}
\label{subsec:training}

\subsubsection{The Noise2Noise approach}
\label{subsubsec:noise2noise}

Noise2Noise was introduced in Ref.~\cite{Lehtinen2018} as a self-supervised learning strategy demonstrating that neural networks can be trained to denoise signals without requiring clean ground-truth data. In simple terms, the network learns to identify the underlying signal by observing which features vary randomly from one acquisition to another and suppresses these variations. The approach relies on pairs of independently noisy measurements of the same underlying signal. When the noise is statistically centered on the true signal and uncorrelated between the paired observations, minimizing the loss between noisy inputs leads, in expectation, to the recovery of the latent clean signal. Under the acquisition conditions considered here, these assumptions are well satisfied to a good approximation. Repeated spectra recorded at the same spatial location share the same underlying Raman signal, while differences between repetitions are primarily driven by stochastic acquisition noise. This noise can be regarded as zero-mean with respect to the signal and exhibits only limited inter-acquisition correlations, except at the longest integration times discussed in \autoref{subsec:noise} \cite{Jahn2021}.

By removing the need for high-quality reference spectra, Noise2Noise reduces experimental overhead and allows training directly from raw, imperfect measurements. This makes the approach particularly suitable for rapid measurements or \textit{in situ} applications, where acquiring long-exposure, low-noise spectra may be impractical \cite{Platt2025}.

To meet the time constraints of the proposed production workflow---where clean acquisitions would be prohibitively expensive for supervised denoising of short-exposure spectra---we adopt a Noise2Noise training scheme. For this purpose, we developed a dedicated implementation tailored to Raman spectral data.

We retained 20 repeated acquisitions per spatial point as a practical compromise that provides sufficient noise diversity for effective Noise2Noise training, while keeping the additional acquisition time low enough to remain compatible with high-throughput workflows.

When more than two independent noisy realizations are available, the Noise2Noise framework naturally extends to dynamic pairing, since any pair of repeated measurements constitutes a valid training sample \cite{Lehtinen2018}. This strategy is consistent with self-supervised denoising approaches that repeatedly sample different observations of the same signal to increase noise diversity and reduce overfitting \cite{Batson2019}.

With this acquisition strategy, the training dataset requires approximately 22\% of the acquisition time of the final high-resolution Raman map. This overhead is explicitly accounted for in the evaluation of the overall speedup enabled by the proposed denoising workflow.

In practice, Noise2Noise can remain effective even when its ideal assumptions are only approximately satisfied. In the present work, this is treated as an empirical approximation supported by the averaging analysis in \autoref{subsec:noise} and by the observed reconstruction performance, rather than as a strict proof that the original Noise2Noise assumptions hold exactly. This robustness is also consistent with later self-supervised extensions, which explicitly considered training pairs that only approximately meet the original Noise2Noise conditions \cite{Zhao2022b,Mansour2023}.

\subsubsection{Training parameters}

Model training was performed using the Adam optimizer \cite{Kingma2017} with a learning rate of 0.0005, a batch size of 128, and training over 500 epochs on a CUDA-enabled device for accelerated computation. To ensure reliable performance estimation and avoid data leakage between training and validation sets, a 5-fold cross-validation scheme \cite{Kohavi1995} was applied, with spectra from the same spatial point grouped together.

\subsection{Classification}

K-means clustering was selected because of its widespread use in Raman hyperspectral analysis, where it serves as a simple and reliable baseline for assessing whether denoising preserves the spectral features relevant for phase discrimination \cite{Abdolghader2021}. The algorithm is applied to the denoised spectra to generate spatial phase maps of the sample (\autoref{fig:kmeans}). As an unsupervised method, K-means groups spectra based on similarity in their spectral profiles, enabling direct evaluation of cluster separability without requiring labelled training data \cite{Ahmed2020}. The number of clusters is determined by combining elbow-curve analysis \cite{Thorndike1953} with visual inspection of spatial and spectral coherence, ensuring that the chosen partition is both statistically motivated and physically interpretable.

\subsection{Evaluation metrics}
\label{subsec:metrics}

The performance of the denoising pipeline is evaluated using a set of complementary quantitative metrics designed to assess both spectral fidelity and the impact of denoising on subsequent analytical steps. At the signal level, three standard metrics are considered. The Root Mean Squared Error (RMSE) measures the average pointwise deviation between a denoised spectrum and its corresponding averaged reference spectrum. It provides a global indicator of reconstruction accuracy and is particularly sensitive to large intensity deviations.

To complement RMSE, the linear Signal-to-Noise Ratio (SNR) is computed as

\[
\mathrm{SNR} = \frac{P_{\mathrm{signal}}}{P_{\mathrm{noise}}},
\]

where $P_{\mathrm{signal}}$ denotes the average power of the reference spectrum, typically defined as the mean squared intensity over all spectral channels, and $P_{\mathrm{noise}}$ denotes the average power of the residual noise, defined in the same way from the difference between the denoised and reference spectra. SNR quantifies the strength of the useful spectral signal relative to the residual noise. Higher SNR values correspond to lower residual noise for a given signal intensity. In Raman spectroscopy, this metric is particularly informative for evaluating the preservation of weak vibrational features after denoising \cite{Hore2010}.

In addition to error-based metrics, the Structural Similarity Index (SSIM) \cite{ZhouWang2004} is used to assess the similarity between denoised and reference spectra in terms of their structural content. Rather than relying on pointwise intensity differences, SSIM compares local patterns of intensity variations by jointly evaluating mean intensity, contrast, and structural agreement. This formulation is well suited to Raman data, where the preservation of peak shapes, relative intensities, and fine spectral features is essential for reliable spectral interpretation.

Beyond spectral fidelity, performance is also evaluated using task-oriented criteria that reflect practical use cases. The effective acquisition speedup enabled by the proposed workflow is estimated by combining the reduced acquisition time associated with short-exposure measurements, the additional time required to acquire the training subset and a recommended test set, and the total computational cost on the specified hardware platform. Finally, clustering accuracy obtained from the K-Means analysis is reported to verify that denoising does not compromise phase discrimination. In \autoref{tab:results}, this metric is computed relative to the clustering obtained from the averaged reference spectra, with cluster correspondence established by the Hungarian algorithm \cite{Kuhn1955} to account for the arbitrariness of K-Means labels. In \autoref{subsec:final-map}, by contrast, the reported agreement is defined relative to the noisy 100~ms map. This metric provides a direct connection between numerical denoising performance and its impact on downstream chemical mapping tasks.

Accordingly, this metric quantifies how closely the clustering structure recovered from the dataset under consideration matches that defined on the averaged-reference dataset, rather than comparing the result with an absolute external label set.

\section{Results}

\subsection{Cross-evaluation}

\begin{figure}[h!]
    \centering
    \includegraphics[width=0.8\linewidth]{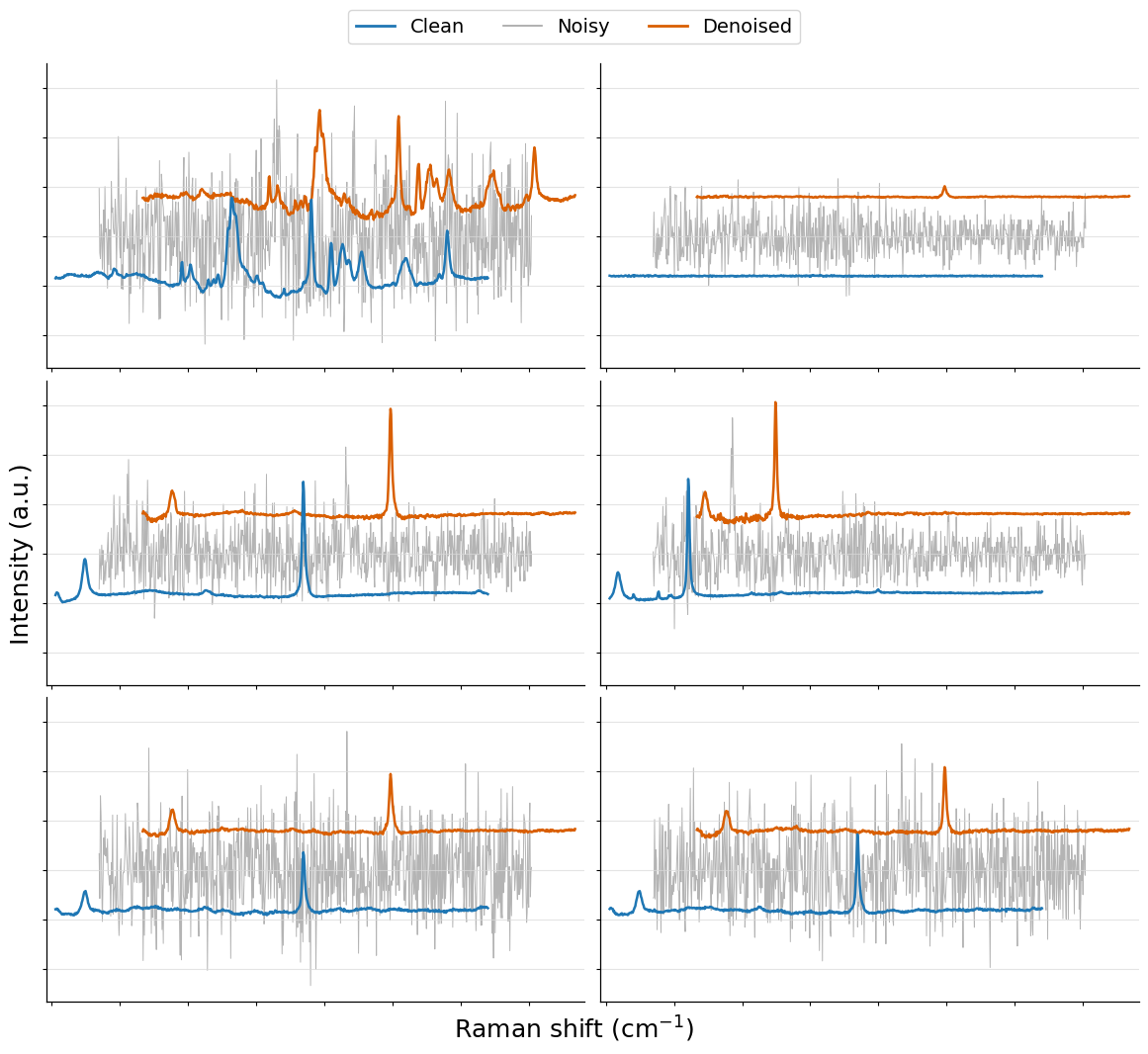}
    \caption{Examples of denoised Raman spectra acquired with a 5~ms integration time. The spectra are displayed with a diagonal offset for better readability. The Noise2Noise reconstruction reduces high-frequency noise while preserving peak shapes and relative intensities, yielding spectra that closely match the averaged reference spectra.}
    \label{fig:accuracy}
\end{figure}

\begin{table}[h!]
\centering
\rowcolors{3}{tablegray}{white}
\setlength{\tabcolsep}{5pt}
\renewcommand{\arraystretch}{1.15}
\resizebox{\linewidth}{!}{%
\begin{tabular}{>{\bfseries}l !{\color{tableline}\vrule width 0.8pt} *{4}{c} !{\color{tableline}\vrule width 0.8pt} *{4}{c}}
\toprule
\rowcolor{tableblue}
& \multicolumn{4}{c!{\color{tableline}\vrule width 0.8pt}}{\textbf{Noisy}} & \multicolumn{4}{c}{\textbf{Denoising Autoencoder}} \\
\rowcolor{tableblue}
Acquisition Time & RMSE & SSIM & SNR & KMeans & RMSE & SSIM & SNR & KMeans \\
\midrule
5 ms & $1.785 \times 10^{-1}$ & 0.0287 & 0.0497 & 95.92\% & $1.072 \times 10^{-2}$ & 0.9192 & 13.7700 & 95.04\% \\
10 ms & $1.153 \times 10^{-1}$ & 0.0871 & 0.1190 & 97.29\% & $9.071 \times 10^{-3}$ & 0.9264 & 19.2300 & 96.78\% \\
25 ms & $6.879 \times 10^{-2}$ & 0.2345 & 0.3344 & 98.40\% & $6.308 \times 10^{-3}$ & 0.9610 & 39.7700 & 98.21\% \\
50 ms & $4.683 \times 10^{-2}$ & 0.3809 & 0.7217 & 98.88\% & $5.835 \times 10^{-3}$ & 0.9661 & 46.4800 & 98.78\% \\
100 ms & $3.581 \times 10^{-2}$ & 0.5070 & 1.2341 & 98.90\% & $5.388 \times 10^{-3}$ & 0.9688 & 54.5200 & 98.78\% \\
5$\times$100 ms & $1.655 \times 10^{-2}$ & 0.7918 & 5.7800 & 99.23\% & $4.424 \times 10^{-3}$ & 0.9775 & 80.8500 & 98.93\% \\
\bottomrule
\end{tabular}%
}

\vspace{1.0em}

\rowcolors{3}{tablegray}{white}
\resizebox{\linewidth}{!}{%
\begin{tabular}{>{\bfseries}l !{\color{tableline}\vrule width 0.8pt} *{4}{c} !{\color{tableline}\vrule width 0.8pt} *{4}{c} !{\color{tableline}\vrule width 0.8pt} *{4}{c}}
\toprule
\rowcolor{tableblue}
& \multicolumn{4}{c!{\color{tableline}\vrule width 0.8pt}}{\textbf{Savitzky-Golay Filter}} & \multicolumn{4}{c!{\color{tableline}\vrule width 0.8pt}}{\textbf{Fourier Transform}} & \multicolumn{4}{c}{\textbf{Wavelet Transform}} \\
\rowcolor{tableblue}
Acquisition Time & RMSE & SSIM & SNR & KMeans & RMSE & SSIM & SNR & KMeans & RMSE & SSIM & SNR & KMeans \\
\midrule
5 ms & $3.790 \times 10^{-2}$ & 0.5123 & 1.1018 & 62.63\% & $3.523 \times 10^{-2}$ & 0.5000 & 1.2754 & 67.34\% & $3.948 \times 10^{-2}$ & 0.4340 & 1.0153 & 71.67\% \\
10 ms & $3.162 \times 10^{-2}$ & 0.5297 & 1.5826 & 96.26\% & $2.929 \times 10^{-2}$ & 0.5675 & 1.8441 & 96.11\% & $3.032 \times 10^{-2}$ & 0.5680 & 1.7220 & 94.79\% \\
25 ms & $2.425 \times 10^{-2}$ & 0.6308 & 2.6915 & 98.34\% & $2.225 \times 10^{-2}$ & 0.6381 & 3.1968 & 98.33\% & $2.242 \times 10^{-2}$ & 0.6454 & 3.1495 & 97.59\% \\
50 ms & $1.874 \times 10^{-2}$ & 0.7008 & 4.5080 & 98.79\% & $1.759 \times 10^{-2}$ & 0.7030 & 5.1122 & 98.82\% & $1.723 \times 10^{-2}$ & 0.7473 & 5.3317 & 98.28\% \\
100 ms & $1.570 \times 10^{-2}$ & 0.7522 & 6.4173 & 98.92\% & $1.479 \times 10^{-2}$ & 0.7841 & 7.2347 & 98.90\% & $1.448 \times 10^{-2}$ & 0.8018 & 7.5488 & 98.47\% \\
5$\times$100 ms & $9.042 \times 10^{-3}$ & 0.8962 & 19.3579 & 99.18\% & $8.773 \times 10^{-3}$ & 0.8952 & 20.5615 & 99.18\% & $8.307 \times 10^{-3}$ & 0.9071 & 22.9311 & 99.40\% \\
\bottomrule
\end{tabular}%
}

\caption{Quantitative comparison of spectral quality and clustering performance for raw and Noise2Noise-denoised spectra at different acquisition times. RMSE, SNR, and SSIM quantify spectral agreement with the averaged reference spectra, while K-Means accuracy in this table evaluates the effect of denoising on phase discrimination by comparison with the clustering obtained from the same averaged reference spectra. The 5$\times$100\,ms spectra were generated by averaging five independent 100\,ms acquisitions. The table also reports results from Savitzky--Golay, Fourier, and wavelet denoising, providing a baseline for comparison with traditional approaches. Details of this benchmark are given in \autoref{subsec:tradition}.}
\label{tab:results}
\end{table}

To obtain a robust estimate of denoising performance, the Noise2Noise evaluation metrics are computed within a $k$-fold cross-validation framework. This strategy reduces sensitivity to a particular data split and provides a more reliable assessment of model generalization in the presence of experimental variability inherent to Raman spectral acquisitions \cite{Kohavi1995,Litjens2017}. The benchmark against classical denoising methods reported in the same table was performed in a separate cross-validation framework, described in \autoref{subsec:tradition}.

In practice, the dataset is partitioned into $k$ disjoint folds, with the constraint that all repeated acquisitions corresponding to the same spatial point are assigned to the same fold, in order to prevent information leakage between training and evaluation. At each iteration, the denoising model is trained on $k-1$ folds and evaluated on the remaining fold. Quantitative metrics are then computed on this held-out subset by comparing both the raw noisy spectra and the Noise2Noise-denoised outputs against the corresponding averaged reference spectra. The reported values correspond to the mean of each metric over the $k$ validation runs. Results are shown in \autoref{tab:results}. Note that the 500\,ms reference spectra (not available in our acquisition protocol) are in fact averages from $N=5$ independent spectra acquired with 100\,ms integration time at the same spatial location. Assuming a stationary signal and independent acquisition-to-acquisition noise, averaging $N$ spectra is equivalent to increasing the integration time by a factor $N$ for shot-noise--dominated contributions, yielding an SNR improvement proportional to $\sqrt{N}$. In contrast, detector read noise is introduced at each acquisition and therefore accumulates when multiple short-exposure spectra are averaged, whereas it is incurred only once in a single long-exposure measurement. Since read noise is typically negligible at long integration times in Raman spectroscopy, the 5$\times$100\,ms averaged spectrum provides a good approximation of a 500\,ms acquisition for shot-noise--limited conditions, but should nevertheless be regarded as a conservative proxy rather than a strict equivalent.

Across all acquisition times, the proposed Noise2Noise autoencoder consistently improves spectral quality relative to the raw noisy data, as evidenced by substantial reductions in RMSE and corresponding increases in SNR and SSIM (\autoref{tab:results}). These improvements are most pronounced at short exposure times (5--10~ms), where stochastic noise dominates the signal. In this regime, RMSE is reduced by more than one order of magnitude, while SNR increases by a factor of about 277 at 5~ms (from 0.0497 to 13.77) and about 162 at 10~ms (from 0.1190 to 19.23), indicating effective suppression of high-frequency noise components.

The SSIM values provide complementary insight beyond purely intensity-based metrics. For 5~ms acquisitions, SSIM increases from near-zero values in the raw data to values exceeding 0.92 after denoising, indicating that the autoencoder restores the structural organization of the spectra, including peak positions and relative shapes, despite the severe degradation of the original measurements. This result shows that the denoising process preserves spectrally meaningful features rather than performing simple signal smoothing.

At longer acquisition times (25--100~ms), where the initial signal-to-noise ratio is higher, the relative gains are naturally reduced but remain systematic. SSIM stabilizes above 0.96, and the residual RMSE approaches the intrinsic variability observed among the averaged reference spectra themselves. This behavior suggests that the denoiser operates close to the noise floor imposed by experimental repeatability, without introducing artificial bias or oversmoothing.

The smooth and monotonic evolution of all metrics as a function of acquisition time further indicates that the model does not rely on a specific noise regime and remains stable across a wide range of exposure conditions. This robustness follows directly from the Noise2Noise training strategy, which exposes the network to realistic noise statistics during learning.

Beyond signal-level fidelity, the impact of denoising on downstream analysis was assessed using K-Means clustering accuracy, by comparing clustering results obtained from denoised spectra with those derived from averaged reference data. Despite substantial spectral modifications at short exposure times, clustering accuracy remains close to 95\% for 5~ms acquisitions, indicating that the denoising process preserves the discriminative spectral features required for reliable mineral phase separation. This establishes a direct link between numerical denoising performance and practical analytical utility.

This cross-evaluation demonstrates that the proposed denoising model provides statistically robust and physically meaningful improvements in Raman spectral quality, particularly in low-signal regimes, without compromising downstream unsupervised classification. The use of $k$-fold cross-validation further supports the reproducibility and general applicability of the approach in high-throughput Raman mapping workflows.

From an experimental perspective, these results indicate a practical relaxation of the acquisition-time constraints traditionally associated with Raman mapping. In particular, exposure times as short as 5~ms per point -- normally insufficient for reliable spectral interpretation -- can be used without compromising phase discrimination or spectral integrity after denoising. This enables the acquisition of dense hyperspectral maps within time frames compatible with routine laboratory workflows, rather than specialized or time-intensive experiments.

Overall, these results highlight a clear trade-off between spectral fidelity and downstream clustering performance. Denoised spectra acquired with an integration time of 5\,ms exhibit significantly improved signal-level metrics (SSIM, RMSE, and SNR) compared with raw 5$\times$100\,ms spectra, indicating that the autoencoder effectively suppresses noise and restores the structural organization of the Raman signal. This demonstrates a raw acquisition speedup on the order of 100 when comparing 5\,ms denoised spectra to 5$\times$100\,ms references. This gain must, however, be weighted by the acquisition cost of the training and test datasets, as well as by the associated computational overhead, as discussed in \autoref{subsec:speedup}. However, this gain in spectral fidelity is accompanied by a slight decrease in K-Means clustering accuracy relative to long-exposure raw data, suggesting that the nonlinear denoising process introduces limited reconstruction distortions that may affect phase separability and spectral interpretability. Importantly, the Raman maps presented in \autoref{subsec:final-map} show that these effects remain minor in practice, as the denoised 5\,ms data still yield spatially coherent and chemically meaningful maps comparable to those obtained from much longer acquisitions.

\subsection{Final map}
\label{subsec:final-map}

\begin{figure}[h!]
    \centering
    \includegraphics[width=\linewidth]{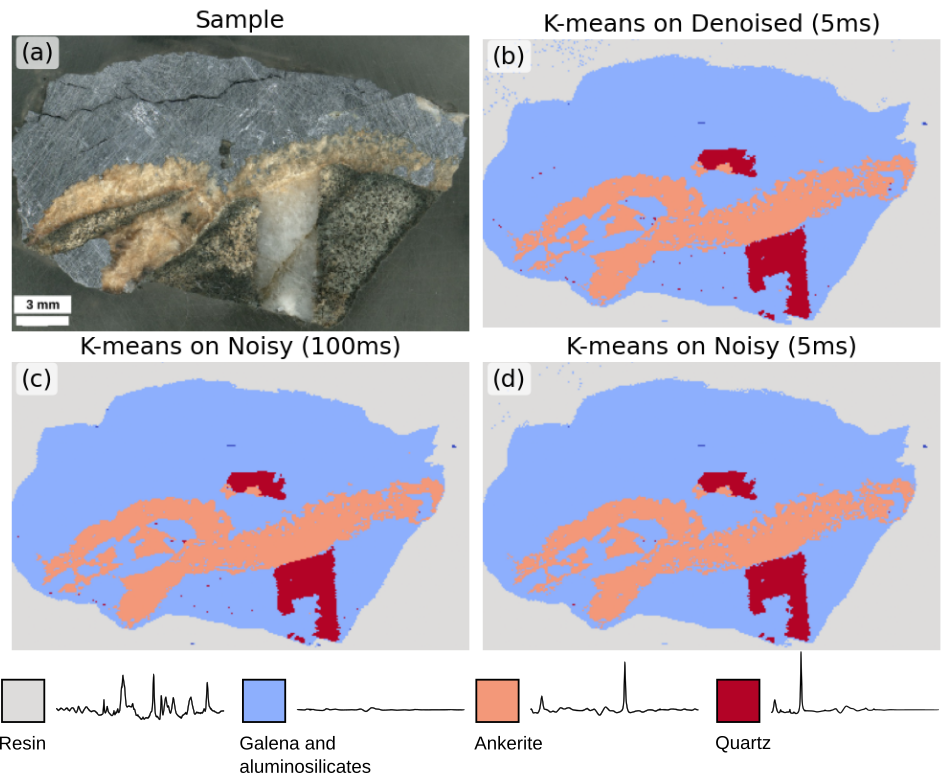}
    \caption{\textbf{Comparison of K-Means clustering results obtained from Raman map under different conditions.} \textbf{(a)} optical image of the composite mineral sample. \textbf{(b)} K-Means clustering applied to the denoised final map: 5 ms integration time. \textbf{(c)} K-Means clustering applied to the noisy map: 100 ms integration time. \textbf{(d)} K-Means clustering applied directly to the noisy map: 5 ms integration time. Representative mean Raman spectra for each cluster are shown below.}
    \label{fig:kmeans}
\end{figure}

\autoref{fig:kmeans} illustrates the effect of the proposed denoising pipeline on the Raman chemical maps obtained by unsupervised K-Means clustering. Clustering performed on denoised spectra acquired with an integration time of only 5 ms per pixel yields spatially coherent phase distributions that closely correspond to those obtained from noisy spectra acquired with a longer integration time of 100 ms.

A quantitative comparison between these two maps shows a 97.97\% agreement in clustering accuracy between the K-Means results derived directly from the 100 ms acquisition and those obtained from the denoised 5 ms data. Unlike the K-Means accuracy reported in \autoref{tab:results}, this value corresponds to a map-to-map agreement with the noisy 100 ms map, not to an agreement with the averaged reference spectra.

Consistent with this high level of agreement, the denoised 5 ms map reproduces well-defined phase boundaries and preserves the spatial continuity of the mineral domains. The representative mean spectra associated with each cluster further indicate that the characteristic Raman signatures are retained after denoising and remain consistent with those extracted from the 100 ms reference acquisition.

Overall, these results indicate that the Noise2Noise-trained autoencoder enables reliable Raman mapping from short-exposure data and remains effective when applied to high-resolution datasets beyond the cross-validation setting. It is worth noting that K-Means clustering performed directly on noisy spectra acquired with a 5 ms integration time (\autoref{fig:kmeans}(d)) already provides a visually consistent representation of the sample, with an accuracy of approximately 96\% as reported in \autoref{tab:results}. This \(\sim\)96\% value is defined relative to the averaged reference spectra, whereas the 97.97\% value reported above is defined relative to the noisy 100 ms final map. As shown in \autoref{tab:results}, denoising leads to a systematic but marginal decrease in K-Means accuracy relative to the averaged-reference clustering. By contrast, the denoised 5 ms final map remains in very close agreement with the noisy 100 ms map. This result highlights an important distinction between spatial mapping and spectral interpretation: the noisy 5~ms spectra remain sufficiently structured for K-Means to recover coherent spatial domains, partly because averaging within clusters attenuates random fluctuations, but the cluster-mean spectra extracted from noisy data remain less readable for mineral attribution than the denoised spectra, especially when weak bands, peak positions, and relative intensities must be inspected directly. The primary interest of the proposed method therefore lies in improving spectral quality for phase identification and interpretation, rather than in enhancing phase mapping itself, which can already be performed efficiently using K-Means clustering on noisy spectra.

\subsection{Method speedup}
\label{subsec:speedup}

\begin{table}[ht]
\centering
\rowcolors{3}{tablegray}{white}
\setlength{\tabcolsep}{6pt}
\renewcommand{\arraystretch}{1.15}
\begin{tabular}{>{\bfseries}l !{\color{tableline}\vrule width 0.8pt} c c c c}
\toprule
\rowcolor{tableblue}
\textbf{Stage} 
& \textbf{Preprocessing} 
& \textbf{Despiking} 
& \textbf{Training} 
& \textbf{Denoising} \\
\midrule
Compute device 
& CPU 
& CPU (16 threads)
& GPU 
& GPU \\

Data 
& Trainset 
& Trainset
& Trainset 
& Final map \\

Total time 
& $<1$ s 
& 15.65 s ($\pm$ 0.19 s) 
& 20.31 s ($\pm$ 0.15 s) 
& 1.12 s ($\pm$ 0.02 s) \\

Time per point 
& $<1.42$ ms 
& 22.2 ms 
& 28.8 ms 
& 0.019 ms \\

Time per spectrum 
& $<0.07$ ms 
& 1.11 ms 
& 1.44 ms 
& 0.019 ms \\
\bottomrule
\end{tabular}
\caption{Computational costs measured on a laptop workstation (22-core CPU, up to 5.1 GHz; NVIDIA RTX 4070 Laptop GPU with 8 GB VRAM; CUDA 12.4; Debian 13 OS). Preprocessing includes baseline correction and normalization, while despiking is reported separately. All runtimes are reported as the mean over 25 runs, with the corresponding standard deviation.}
\label{tab:computational-cost}
\end{table}

Rather than summarizing performance through a single abstract speedup factor, the duration of a realistic high-throughput Raman mapping workflow is decomposed into its main temporal contributions and compared to a conventional acquisition strategy.

In this study, the final Raman map comprises $64,000$ spectra ($320 \times 200$). Using an integration time of 5 ms per spectrum, the raw acquisition time of the final map is $64,000 \times 5$ ms, corresponding to 320~s (5~min~20~s). As shown above, denoising preserves spectral fidelity and phase discrimination at this minimal exposure time, enabling dense Raman mapping with short integrations.

For comparison, a conventional acquisition aiming at comparable spectral quality would require an integration time of 500 ms per spectrum. In this case, the acquisition of the same map would take $64,000 \times 500$ ms, corresponding to approximately 8~h 53~min. The short-exposure denoising strategy therefore reduces the raw acquisition time of the final map by a factor of 100.

The proposed production workflow additionally includes the acquisition of a training set and a test (validation) set. For the 5~ms production estimate, the training subset contains 704 spatial points and 20 repeated acquisitions per point, giving $704 \times 20 \times 5$~ms = 70.4~s, i.e. about 1~min~10~s, or approximately 22\% of the raw acquisition time of the final map. The validation set is counted separately and is estimated from about 60 spatial points, corresponding to approximately 10\% of the training set, acquired with ten 100~ms repetitions per point; this gives about 60~s of additional acquisition time. Importantly, these overheads are bounded and do not scale with the size or resolution of the final map.

The computational costs reported in \autoref{tab:computational-cost} were measured on a standard laptop to emphasize the practical deployability of the proposed pipeline. According to \autoref{tab:computational-cost}, despiking requires 15.65~s and training 20.31~s, corresponding to a combined computational overhead of 36~s. Relative to the acquisition-only duration of the proposed workflow (about 7~min~30~s), this represents approximately 8\%. These costs are therefore not negligible. This makes a lightweight design essential. Once trained, the denoising step itself introduces only 1.12~s of additional latency for the full final map, so inference remains negligible.

Adding the final-map, training, and validation acquisitions gives an acquisition-only duration of about $320+70+60 \simeq 450$~s, i.e. 7~min~30~s. Including these additional acquisitions and the 36~s computational overhead, the total workflow duration required by the proposed pipeline is approximately 8~min~7~s, compared with approximately 9~h for a conventional long-exposure mapping strategy. This corresponds to an effective speedup of approximately a factor of 65 in total workflow time.

Overall, the acceleration provided by the proposed approach primarily arises from the ability to perform reliable denoising of short-exposure spectra. While additional acquisition time is required for training and validation data, this overhead remains modest relative to the gain obtained on the final map. The resulting workflow therefore offers a practical and quantitatively significant improvement in measurement efficiency while maintaining spectral and analytical reliability.

\subsection{Comparison with classical denoising methods}
\label{subsec:tradition}

We compared the proposed denoising autoencoder with three widely used classical denoising methods: Savitzky--Golay smoothing \cite{Savitzky1964}, Fourier filtering, and wavelet thresholding \cite{Mallat1999}. This benchmark was performed in a separate cross-validation framework from the Noise2Noise model evaluation, because these classical filters do not involve neural-network training but do require method-specific hyperparameter selection. For each acquisition time and validation split, the hyperparameters of each method were optimized on a random subset of the training portion by minimizing the RMSE, and the selected configuration was then evaluated on the corresponding held-out spectra. These hyperparameters included the window length and polynomial order for Savitzky--Golay smoothing, the cutoff frequency and filter shape for Fourier filtering, and the wavelet family, decomposition level, and thresholding strategy for wavelet denoising. The values reported in \autoref{tab:results} correspond to the mean over these classical-method validation splits. This protocol enables a fair comparison between noisy spectra, classical denoising strategies, and the proposed autoencoder while keeping the validation procedure adapted to each method family.

The results reported in \autoref{tab:results} show that the present autoencoder consistently provides the best overall performance across all acquisition times, with the largest advantage in the short-exposure regime that is most relevant for high-throughput Raman measurements. At 5~ms, it achieves the lowest RMSE ($1.072 \times 10^{-2}$), clearly outperforming Savitzky--Golay ($3.790 \times 10^{-2}$), Fourier filtering ($3.523 \times 10^{-2}$), and wavelet denoising ($3.948 \times 10^{-2}$). The same trend is observed for SSIM and SNR: the autoencoder reaches 0.9192 and 13.77, respectively, whereas the classical methods remain limited to SSIM values of 0.43--0.51 and SNR values close to 1. These results indicate that conventional filters can smooth the spectra, but are less effective at restoring the structural features of the Raman signal.

This difference is also reflected in downstream clustering performance. At 5~ms, the proposed method preserves a K-Means accuracy of 95.04\%, compared with 62.63\% for Savitzky--Golay, 67.34\% for Fourier filtering, and 71.67\% for wavelet denoising. At longer acquisition times, the performance gap becomes smaller as the input spectra are less degraded, and all methods converge toward similarly high clustering accuracies. Overall, these results show that the main advantage of the proposed approach lies in its ability to preserve both spectral fidelity and analytical usefulness in the most noise-limited acquisition conditions.

\vspace{0.75\baselineskip}
\section{Discussion and Conclusion}

The results presented in this study enable the definition of a practical production workflow for high-throughput Raman spectroscopy based on short-exposure acquisitions combined with a Noise2Noise denoising strategy. The proposed approach is designed to substantially reduce, by a factor of about 65 in the present study, acquisition time while preserving spectral fidelity and phase discrimination, and to remain compatible with routine laboratory constraints.

The workflow begins with the acquisition of a reduced training subset over a limited number of spatial points selected to be representative of the phases present in the sample. At each of these points, several short-exposure repetitions are recorded and used to train a fully convolutional denoising autoencoder following the Noise2Noise paradigm. Because this strategy does not require clean high-SNR reference spectra for training, it avoids the need for long acquisitions during training and remains compatible with realistic experimental conditions. The acquisition time associated with this training subset represents only a bounded fraction of the total mapping time.

A compact validation subset, acquired at a small number of additional locations with a higher number of repetitions per point, provides high signal-to-noise reference spectra for performance verification. This validation step offers a practical means to monitor denoising quality and detect potential reconstruction artefacts during training, without significantly increasing the overall experimental overhead.

Once trained, the autoencoder can be applied to the final Raman map, which is acquired using short integration times per pixel. The denoised spectra can be directly used for downstream analysis, such as unsupervised clustering, yielding chemical maps comparable to those obtained from much longer acquisition times. More importantly, denoising restores a level of spectral readability that facilitates phase attribution and analyses performed at the individual-spectrum level. Even when clustering remains adequate, short-exposure raw spectra are still difficult to interpret directly. By improving peak visibility and the overall spectral profile, denoising facilitates comparison with reference databases and is expected to support subsequent operations such as peak fitting, spectral decomposition, or compositional interpretation. Because both preprocessing and inference have a computational cost that remains small compared with typical Raman integration times, the denoising pipeline can, in principle, be applied during acquisition. This opens the perspective of near real-time visualization or analysis of denoised spectra in future implementations.

Quantitative evaluation confirms that the proposed pipeline substantially improves spectral fidelity, particularly at very short exposure times where stochastic noise dominates the signal. Improvements in RMSE, SNR, and SSIM demonstrate that the denoiser effectively suppresses noise while preserving peak positions, shapes, and relative intensities. At the same time, K-Means clustering applied directly to noisy spectra remains robust even at 5~ms (about 96\% accuracy). In this context, the main contribution of denoising is not to recover separability \textit{per se}, but to preserve it while making the spectra substantially more interpretable.

The workflow can be retrained for each new sample, making it well suited for adaptable and routine high-throughput Raman measurements. For the static-sample mapping considered here, building the training set is straightforward and represents about 22\% of the acquisition time of the final map. Compared with supervised denoising, Noise2Noise also remains experimentally less restrictive because it only requires repeated short-exposure measurements of the same sample. Beyond the specific Raman case study presented here, the proposed approach constitutes a general and transferable framework for the denoising and analysis of one-dimensional spectroscopic signals. The combination of standardized preprocessing, a lightweight fully convolutional architecture, and a Noise2Noise training strategy relies on limited modality-specific assumptions and may therefore be applicable to other spectroscopic techniques characterized by localized spectral features and stochastic noise, such as infrared absorption or photoluminescence spectroscopy.

Overall, this work demonstrates that reliable denoising of short-exposure spectra can be achieved without sacrificing analytical performance, thereby relaxing traditional acquisition-time constraints in Raman hyperspectral imaging. The proposed pipeline provides a practical basis for accelerated spectroscopic mapping in the static-sample setting considered here, and it may also be relevant for Raman measurements performed without spatial redundancy, such as single-point acquisitions. A natural perspective of this work is online or real-time Raman analysis, although this extends beyond the static acquisition setting studied here. Time-resolved measurements would require either repeated states or adapted strategies, such as pre-training, transfer learning, or incremental training.

Finally, we expect that the present approach can be extended to a broad range of high-throughput spectroscopic analyses. Large-scale analytical techniques such as photoluminescence mapping (e.g., confocal microscopy) and elemental mapping by laser-induced breakdown spectroscopy (LIBS) could constitute promising application domains for this pipeline, provided that their noise statistics and acquisition constraints remain compatible with a Noise2Noise strategy, in particular the availability of repeated measurements with approximately independent noise.

\medskip
\textbf{Data Availability Statement} \par
The experimental Raman datasets supporting the findings of this study are openly available on Zenodo (DOI: \href{https://doi.org/10.5281/zenodo.18244161}{10.5281/zenodo.18244161}). The repository contains raw Raman acquisitions at multiple integration times, used for training, validation, and evaluation of the Noise2Noise denoising pipeline.

\medskip
\textbf{Code Availability Statement} \par
The source code for the production pipeline is openly available on Zenodo and GitHub (DOI: \href{https://doi.org/10.5281/zenodo.18154207}{10.5281/zenodo.18154207}).

\medskip
\textbf{Acknowledgements} \par 
The authors thank Jean Cauzid and Cécile Fabre (Université de Lorraine, GeoRessources Laboratory, UMR CNRS 7359, France) for providing the mineral sample used in this study. They also thank Serge Buathier for his valuable contribution to the setup of the Raman instrumentation. The authors thank Nicola Vigano for valuable discussions about the Noise2Noise approach and Dylan Bissuel for discussions on the autoencoder solution. This work was supported by the French government ``France 2030'' initiative, under the DIADEM program managed by the ``Agence Nationale de la Recherche'', (ANR-22-PEXD-0014, ``Libelul'') and (ANR-22-PEXD-0015 ``Diamond'').

\medskip
\textbf{Conflict of Interest} \par
The authors declare no conflict of interest.


\end{document}